\documentclass{article}
\pdfoutput=1
\usepackage{graphicx}
\usepackage{named}
\usepackage{amsmath}
\usepackage{amsthm}
\usepackage{tabularx,booktabs}
\newcolumntype{Y}{>{\centering\arraybackslash}X}
\newcolumntype{Z}{>{\raggedleft\arraybackslash}X}
\newcolumntype{S}{>{\hsize=.5\hsize\centering\arraybackslash}X}

\usepackage{algorithm}
\usepackage{algorithmic}
\usepackage[colorlinks=true,linkcolor=black,anchorcolor=black,citecolor=black,filecolor=black,menucolor=black,runcolor=black,urlcolor=black]{hyperref}
\usepackage{url}
\usepackage{amssymb}
\usepackage{xspace}
\usepackage{color}
\newcommand{\placefigure}[2]{%
  \vfill
  \begin{center}
    \includegraphics[scale=#1]{#2}
  \end{center}
  \vfill
}
\hypersetup{pdfauthor={Daniel Borrajo, Manuela Veloso and Sameena Shah},pdftitle={Simulating and Classifying Behavior in Adversarial Environments Based on Action-State Traces: An Application to Money Laundering}}

\newcommand{\no}[1]{}
\newcommand{\change}[1]{#1}
\newcommand{\myeq}[1]{%
  \vspace*{0.25cm}
  \centerline{#1}
  \vspace*{0.25cm}
}
\newcommand{\cabbot}{{\sc cabbot}\xspace}
\newcommand{\ribl}{{\sc ribl}\xspace}

\newcommand{\planning}{C}
\newcommand{\observer}{F}

\title{Simulating and Classifying Behavior in Adversarial Environments Based on Action-State Traces: An Application to Money Laundering} 

\author{Daniel Borrajo\thanks{On leave from Universidad Carlos III de Madrid. The position at the lab is as a consultant.}
  \and Manuela Veloso\thanks{On leave from Carnegie Mellon University} \and Sameena Shah\\
J.P.Morgan AI Research, New York, NY, USA\\
\{daniel.borrajo, manuela.veloso, sameena.shah\}@jpmchase.com}

\begin{document}

\maketitle

\begin{abstract}
Many business applications involve adversarial relationships in which both sides adapt their strategies to optimize their opposing benefits. One of the key characteristics of these applications is the wide range of strategies that an adversary may choose as they adapt their strategy dynamically to sustain benefits and evade authorities. In this paper, we present a novel way of approaching these types of applications, in particular in the context of Anti-Money Laundering. We provide a mechanism through which diverse, realistic and new unobserved behavior may be generated to discover potential {\it unobserved adversarial actions} to enable organizations to preemptively mitigate these risks. In this regard, we make three main contributions. (a) Propose a novel behavior-based model as opposed to individual transactions-based models currently used by financial institutions. We introduce {\it behavior traces} as enriched relational representation to represent observed human behavior. (b) A modelling approach that observes these traces and is able to accurately infer the goals of actors by classifying the behavior into money laundering or standard behavior despite significant unobserved activity. And (c) a synthetic behavior simulator that can generate new previously unseen traces. The simulator incorporates a high level of flexibility in the behavioral parameters so that we can challenge the detection algorithm. Finally, we provide experimental results that show that the learning module (automated investigator) that has only partial observability can still successfully infer the type of behavior, and thus the simulated goals, followed by customers based on traces - a key aspiration for many applications today.

\end{abstract}

\section{Introduction}
 Adversarial settings are common in many business domains, where both sides adapt their strategies over time. For example, money laundering vs Anti-Money Laundering (AML), or those in fraud, and cyber-crime. If we take AML, a key characteristic is the wide range of strategies that are available to a money launderer, who may come up with completely novel, previously unseen strategies to evade authorities. At present, models available to investigators involve playing catch-up. Investigators may happen to detect a new typology used by a money launderer and may make recommendations to put in new controls for the novel typology detected. Subsequently, the organization may adjust their models to counter the newly observed strategy to allow its detection going forward. However, significant delay and crime might have passed through by the time new controls are put in place. Money launderers, fraudsters, or other bad actors often remove and obscure the funds (benefits) from the networks to nullify their risk of any future cease of funds, in case of any retrospective action by authorities. Hence, timely detection of previously unseen typologies is of utmost importance. 
 
In this paper, we focus on AML. As defined by Senator et al.~(\citeyear{Senator95}), ``Money laundering is a complex process of placing the profit \ldots from illicit activity into the legitimate financial system, with the intent of obscuring the source, ownership, or use of the funds.'' 
Over time, financial institutions have been mandated by law enforcement agencies to improve their processes to detect
suspicious activity and raise the corresponding Suspicious Activity Reports (SARs). A typical prevalent AML model starts
by observing transactions, public media, or a referral, and generates alerts~\cite{isa2015}. Then, alerts are
investigated by humans who decide whether they need to report a SAR to law enforcement for the alert. Since there is a
bias towards being conservative and raising alerts at the detection of any suspicious behavior, many alerts are
generated and the manual effort put into investigations is enormous~\cite{takats2011theory}.  However, despite all these
efforts, most of the money laundering activities are not noticed in time. Europol estimates that less than 1\% of money
originating from criminal activities is recovered in Europe.\footnote{\url{https://www.europol.europa.eu/publications-documents/does-crime-still-pay}, accessed May 27, 2020}

In order to provide efficient AI tools to help human investigators and law enforcement, 
any investigation needs to provide rationale for its decisions, as filing a SAR.\footnote{\url{https://www.fincen.gov/sites/default/files/shared/CTRPamphlet.pdf}, accessed May 26, 2020.}
Rationale usually constitutes of gathering all ``applicable'' set of evidences via a sequence of
steps and a final assessment of all evidences. Mostly, applicable evidences deal with inferring subject's goals.  
Therefore, in order to be
used in practice, the output of any AI-based system that tackles this task should explicitly mention the goals the
suspects were pursuing, as well as the actions the subjects of interest carried out and the evidences that led to that
conclusion.

Previous work on automating AML has defined sets of rules to detect particular behaviors~\cite{Rajput14,Senator95}, or applied machine learning to transactions or social networks~\cite{Chen18,colladon2017}. Most of these recent works use neural network (NN) based approaches~\cite{weber2018scalable,schlichtkrull2018modeling}. These approaches have shown exceptional performance for
many tasks including image classification or natural language processing. But, given their current way of handling
explanations, they are not yet a viable solution from a practical perspective. Instead, symbolic approaches are clearly
much better suited for tasks requiring explanations.
Other approaches used a variety of AI techniques to solve the AML classification task, such as SVMs~\cite{Kingdon04},
Dynamic Bayesian Networks and clustering~\cite{Raza11}, RBF neural networks~\cite{Lin-Tao08}, fuzzy logic~\cite{Yu-To11}, association rules and frequent set analysis~\cite{Luo14},
clustering~\cite{Larik2011} or decision trees~\cite{Su-Nan07}. In general, most of these papers and others published on applying machine learning (or
other AI techniques) to AML, do not include comparison with other works, and use their own simulators and data. We include in this paper a comparison of our relational learning technique against a decision tree.  The AML task can also be posed as anomalies or outliers
  detection~\cite{chandola2010anomaly,gupta2013outlier} where the techniques and representation formalisms are also based on attribute-value.

In this paper, we assume there is a rationale for the behavior of agents (customers) when taking actions in the
environment that is partially observable by a financial institution. This behavior depends on some hidden human goals and
the states they encounter while taking actions to achieve those goals. We also assume goals, states and
actions can be represented using a form of high-level representation formalism, such as predicate logic. These assumptions are in line with the need to file rationale for each SAR. The evidences compiled in SARs correspond to descriptions of actions (activities) taken by suspicious persons (e.g. several cash deposits) and the corresponding states (e.g. network of people and companies). While all this knowledge could potentially be represented in the attribute-value representation used by most other AI techniques applied to AML, usually we have to constrain the size of the representation, require some extensive domain-dependent feature engineering or lose representation power~\cite{dzeroski2010relational}.

Given those assumptions, we can pose the problem of AML as a relational classification
task~\cite{dzeroski2010relational}. It takes as input a trace of human behavior corresponding to the execution of
observable actions by the financial institution and the corresponding observable components of states. It generates as output
a decision of whether that trace corresponds to money laundering or not. We train
our learning system with previous traces of known behavior (both money laundering or not), which can be trivially extracted from current information systems of
financial institutions in a relational format.
We name our learning system \cabbot for Classification of Agents' Behavior Based on Observation Traces. Process mining is a related field whose goals include identifying processes from traces~\cite{ProcessMining2016}. However, most work on process mining assumes actions are represented as labels (no parameters), and there is very little reasoning on states and goals as logical formulae.

We present as main contributions of this research: a new enhanced model of human behavior in the context of financial institutions based on states and actions; a learning technique that can classify in agents' (or behavior) types based on observation traces; and a simulator of agents' behavior based on dynamic goal generation, planning and execution. Since most available datasets on AML correspond to only transactions data, we have built a simulator that generates realistic traces of these kinds of behavior and incorporates a rich representation of actions and
states.

\section{States and Actions Traces}

We assume agents' rational behavior to be based on the concepts of goals, states and actions. In order to establish a
common representation language, we will use a form of predicate logic to represent the information that an agent (such as a financial institution), $\observer$, can observe from the behavior of another agent (e.g. customer), $\planning$. States will be represented as sets of literals, where literals can be predicates or functions. Predicates have a name and a list of arguments (e.g. {\tt account-owner(c1,acc1)}). Functions represent numeric variables and are composed of a name, its arguments and the value (e.g. {\tt balance(acc1)=12020}). As in the real world, part of the state will be observable by $\observer$ and another part will not be observable. For instance, the owner of an account will be observable by a financial institution, while the fact that someone is trying to launder money will not be observable.

\subsection{Representation of States and Goals}
Tables~\ref{tab:predicates} and~\ref{tab:functions} lists the key predicates and functions we have defined. They
can refer to four categories of information: {\it transaction}-based; relationship-based, related to the {\it network} of people or companies connected to each customer; {\it both} kinds, transaction and network; or to the {\it bank}, but not related to network or transactions. The tables also represent their observability. \no{Alice suggests giving an example. No space}

\begin{table}[hbtp]
  \begin{tabularx}{\textwidth}{*{3}{Y}}
    \toprule
    \multicolumn{1}{c}{\bf Predicate} & \multicolumn{1}{c}{\bf Observable} & \multicolumn{1}{c}{\bf Type}\\ \midrule
    money-laundering & no & \\
    money-laundered & no &\\
    has-dirty-money & no & \\
    criminal & no & \\
    banned-country & yes & network\\
    account-owner & yes & network\\
    account-country & yes & network\\
    member-of & yes & network\\
    bill-due & yes & network\\
    owes-money & no &\\
    employed & yes & network\\
    works-for & yes & network\\
    has-company & yes & network\\
    transaction-origin & yes & transactions\\
    transaction-destination  & yes & transactions\\
    received-payroll & yes & both\\
    has-card & yes & bank\\
    enjoyed-service & no & \\
    provides-service& no & \\
    owns& no & \\ \bottomrule
  \end{tabularx}
  \caption{List of predicates, their observability and the type of information they refer to in case they are
    observable.}
  \label{tab:predicates}
\end{table}

\begin{table}[hbtp]
  \begin{center}
  \begin{tabularx}{\textwidth}{*{3}{Y}}
    \toprule
    \multicolumn{1}{c}{\bf Function} & \multicolumn{1}{c}{\bf Observable} & \multicolumn{1}{c}{\bf Type}\\ \midrule
    balance & yes & both\\
    transaction-amount & yes & transactions\\
    dirty-money & no & \\
    criminal-income & no & \\
    working-day & no & \\
    days-without-pay & no & \\
    salary & no & \\
    price & no & \\
    owed-money & no & \\ \bottomrule
  \end{tabularx}
  \end{center}
  \caption{List of functions, their observability and the type of information they refer to in case
    they are observable.}
  \label{tab:functions}
\end{table}

\subsection{Representation of Actions}
Table~\ref{tab:actions} lists the key actions we have defined, together with the same data as in the case of
predicates. We duplicate some of these actions, since they can be used by standard customers or criminals. $\observer$ will observe criminals' actions as their corresponding standard ones. For instance, {\tt
  integration-cash-out} represents a withdrawal following money laundering, but
$\observer$ will observe it as a {\tt cash-out} action.

  \begin{table}[hbtp]
    \begin{center}
  \begin{tabularx}{\textwidth}{Y *{2}{S}}
    \toprule
      \multicolumn{1}{c}{\bf Action} & \multicolumn{1}{c}{\bf Observable} & \multicolumn{1}{c}{\bf Type}\\ \midrule
      create-company & no & \\
      associate & no & \\
      create-account & yes & network\\
      set-ownership-account & yes & network\\
      perform-criminal-action & no & \\
      finish-money-laundering & no & \\
      takes-job & no & \\
      work & no & \\
      payroll & yes & both \\
      quick-deposit & yes & transactions \\
      placement-cash-in & yes & transactions \\
      digital-deposit & yes & transactions \\
      placement-digital & yes & transactions \\
      buy-digital & no & \\ 
      cash-out & yes & transactions \\
      integration-cash-out & yes & transactions \\
      pay-bill & yes & both \\
      create-bill & no & \\
      integration-pay-bill & yes & both \\
      move-funds & yes & transactions \\
      move-funds-internationally & yes & transactions \\
      move-funds-self & yes & transactions \\
      layering & no & \\
      quick-payment & yes & transactions \\
      buy-direct & yes & transactions \\
      placement-buy-direct & yes & transactions \\
      enjoy-service & yes & transactions \\
      placement-enjoyed-service & yes & transactions \\ \bottomrule
    \end{tabularx}
\end{center}
    \caption{List of actions, their observability and the type of information they refer to in case they are
      observable.}
    \label{tab:actions}
  \end{table}

\subsection{Traces of Behavior}
The learning system takes as input traces of observable behavior.  A trace $t_\planning$ is a sequence of states and
actions executed by $\planning$ in those states:
$t_\planning=(s_{0},a_{1},s_{1},a_{2},s_{2},\ldots,s_{n-1},a_{n},s_{n})$,
where $s_{i}$ is a state and $a_{i}$ is an action name and its parameters. States and actions correspond to the observable predicates and actions from the viewpoint of  $\observer$.
We will call $T_\planning=\{t_\planning\}$ the set of traces observed from $\planning$.

We assume there is nothing in the observable state that directly identifies one or the other type of behavior. There
is also no difference on the observable actions between those that can be executed by one or the other type of
$\planning$.

\section{Learning to Classify Behavior}

$\observer$'s task consists of learning to classify among the different types of $\planning$ (behaviors). 

\subsection{Learning Task}
The learning task can be defined as follows.  \textbf{Given}: $N$ classes of behavior, (\{good, bad\} in our current
application);\footnote{Good refers to standard customers' behavior and bad corresponds to money laundering-related
  behavior.} and a set of labeled observed traces,  $T_{\planning_i}, \forall \planning_i\in \{{\mbox good, bad}\}$
\textbf{Obtain}: a classifier that takes as input a new (partial) trace $t$ (with unknown class) and outputs the
predicted class.

A characteristic of this learning task is that it works on unbounded size of the learning examples. Traces can be
arbitrarily large. Also, states within the trace and action descriptions can be
arbitrarily large (both in the number of different action
schemas, and in the number of grounded actions). Using fixed-sized input learning techniques
can be difficult in these cases and some assumptions are made to handle that characteristic. Therefore, we will consider
 only relational learning techniques, and, in particular, relational instance-based approaches.

We use relational $k$NN (\ribl) to classify a new trace according to the $k$ traces with minimum distance, and then computing the mode of those traces' classes. Since the classifier takes a trace as input, \cabbot also allows for on-line classification with the current trace up to a given execution step. A nice property of $k$NN is that we can explain how a behavior was classified by pointing out the closest previous cases and which are the most similar components (actions and states) of the closest traces.

\subsection{Distance Functions between Traces}

The key parameter of the relational learning techniques we will use is the distance between two traces,
$d: T\times T\rightarrow \mathbb{R}$. Similarity functions have been extensively studied~\cite{ontanon2020overview}. We have defined four distance metrics that can deal with state-action traces: actions; state differences; n-grams; and relational.

\textbf{Distance based on Actions.} 
A simple, yet effective, distance function consists of using the inverse of the Jaccard similarity function~\cite{Jaccard} as:

    \myeq{$d_a(t_1,t_2)=1-\frac{|an(t_1)\cap an(t_2)|}{|an(t_1)\cup an(t_2)|}$}

where $an(t_i)$ is the set of actions names in $t_i$. This distance is based on the ratio of common action names in both
traces to the total number of different action names in both traces.

\textbf{Distance based on State Differences.} 
Given two consecutive states $s_1$ and $s_2$ in a
  trace, we define their associated difference or delta as $\delta_{s_i,s_{i+1}}=s_{i+1}\setminus s_i$. These deltas represent
  the new literals in the state after applying the action.
  We can compute a distance between the sets of deltas on each trace by using the Jaccard similarity function as before.

  \myeq{$d_\Delta(t_1,t_2)=1-\frac{|\Delta(t_1)\cap \Delta(t_2)|}{|\Delta(t_1)\cup \Delta(t_2)|}$}

  where $\Delta(t_i)$ is the set of deltas of a trace $t_i$. Again, we only use the predicate and function names.
  
\textbf{Distance based on $n$-grams.}
The two previous distances only consider actions and deltas as sets. If we want to improve the distance metric, we
  can use a frequency-based approach (equivalent to an $n$-grams analysis with $n=1$). Each trace is represented by a
  vector. Each position of the vector contains the number of times an observable action appears in the trace. The
  distance between two traces, $d_g$, is defined as the squared Euclidean distance of the vectors representing the
  traces. As before, a new trace is classified as the class of the training trace with the minimum distance to the new
  trace.

\textbf{Relational-based Distance.}
Instead of using only counts, the distance function can also consider action and state changes as relational
  formulae and use more powerful relational distance metrics. We define a version of the \ribl relational distance
  function~\cite{ribl} adapted to our representation of traces, $d_r$. We modify it given the different
  semantics of the elements of the traces with respect to generic \ribl representation of examples. Given two traces, we
  first normalize the traces by substitution of the constants' names by an index of the first time they appeared
  within a trace.  For instance, given the following action and state:

\begin{center}
\begin{minipage}{\textwidth}
    \begin{tabbing}
      $\langle$ \={\tt create-account(customer-234,acc-345)},\\
      \> {\tt \{}\={\tt acc-owner(customer-234,acc-345)}\\
      \> {\tt balance(acc-345)=2000\}} $\rangle$
    \end{tabbing}
    \end{minipage}
\end{center}

    the normalization process would convert the trace to:

 \begin{center}
\begin{minipage}{\textwidth}
     \begin{tabbing}
       $\langle$ \={\tt create-account(i1,i2)},\\
       \> {\tt \{acc-owner(i1,i2),balance(i2)=2000\}} $\rangle$
     \end{tabbing}
   \end{minipage}
\end{center}

    This process allows the distance metric to partially remove the issue related to using different constant names in
    the traces. The distance $d_r$ is then computed as:  \[d_r(t_1,t_2)=\frac{1}{2}(d_{ra}(t_1,t_2)+d_{r\Delta}(t_1,t_2))\]

    i.e. as the average of the sum of $d_{ra}$ (distance between the actions of the two traces) and $d_{r\Delta}$
    (distance between the deltas of both traces). $d_{ra}$ is computed as:
    \[d_{ra}(t_1,t_2)=\frac{1}{Z} \sum_{a_i\in a(t_1)} \min_{a_j\in a(t_2)} d_{f}(a_i,a_j)\]
  
  where $a(t_i)$ is the set of ground actions in $t_i$, $d_f$ is the distance between two relational formulae and $Z$ is
  a normalization factor ($Z=\max\{|a(t_1)|,|a(t_2)|\}$). We normalize by using the length of the longest set of actions
  to obtain a value that does not depend on the number of actions on each set, so distances are always between 0 and
  1. $d_f$ is 1 if the names of $a_i$ and $a_j$ differ. Otherwise, it is computed as:

  \myeq{$d_{f}(a_i,a_j)= 0.5-0.5\frac{1}{|\mbox{arg}(a_i)|} d_{\mbox{arg}}(a_i,a_j)$}

  where $d_{\mbox{arg}}(a_i,a_j)$ is the sum of the distances between the arguments in the same positions in both
  actions. Each distance will be 0 if they are the same constant and 1 otherwise. Again, we normalize the values for
  distances. Also, when two grounded actions have the same action name, we set a distance of at most 0.5. For instance,
  if we have two literals $l_1=${\tt create-account(i1,i2)} and $l_2=${\tt create-account(i3,i2)},
  
  \myeq{$d_f(l_1,l_2)=0.5-0.5\frac{1}{2}(1+0)=0.25$.}

  As a reminder, each trace contains a sequence of sets of literals that correspond to the delta of two
  states. Therefore, $d_{r\Delta}$ is computed as the distance of two sets of deltas of literals ($\Delta(t_1)$ and
  $\Delta(t_2)$). We use a similar way to compute it as in the previous formulas:

  \myeq{$d_{r\Delta}(t_1,t_2)=\frac{1}{Z_\Delta}\sum_{\delta_1\in\Delta(t_1)} \min_{\delta_2\in\Delta(t_2)} d_{r\delta}(\delta_1,\delta_2)$}

  where $Z_\Delta=\max\{|\Delta(t_1)|,|\Delta(t_2)|\}$, and $d_{r\delta}$:
  \[d_{r\delta}(\delta_1,\delta_2)=\frac{1}{\max\{|\delta_1|,|\delta_2)|\}} \sum_{l_i\in \delta_1} \min_{l_j\in \delta_2} d_{f'}(l_i,l_j)\]
  
  $d_{f'}(l_i,l_j)=d_{f}(l_i,l_j)$ when the literals correspond to predicates. We use $d_f$ since actions and literals
    in the state ($l_j, l_j$) share the same format (a name and some arguments). However, when they correspond to
    functions, since functions have numerical values, we have to use a different function $d_n$. In this case, each
    $l_i$ will have the form $f_i(\mbox{arg}_i)=v_i$. $f(\mbox{arg}_i)$ has the same format as a predicate (or action)
    with a name $f_i$ and a set of arguments $\mbox{arg}_i$, so we can use $d_f$ on that part. The second part is the
    functions' value. In this case, we compute the absolute value of the difference between the numerical values of both
    functions and divide by the maximum possible difference ($M$) to normalize:\footnote{We use a large constant in practice.}

    \myeq{$d_n(l_i,l_j)= d_f(f_i(\mbox{arg}_i),f_j(\mbox{arg}_j))\times \frac{\mbox{abs}(v_i-v_j)}{M}$}

    We multiply both formulae, since we see the distance on the arguments as a weight that modifies the difference in
    numerical values. For example, if

    $\delta_1$={\tt \{acc-owner(i1,i2),balance(i2)=20\}},\\
    \indent $\delta_2$={\tt \{acc-owner(i1,i3),balance(i3)=10\}},
    \[d_{r\Delta}(\delta_1,\delta_2)=\frac{1}{2}(\min\{0.25,1\}+\min\{1,0.5\times \frac{|20-10|}{M}\})\]

\section{Experiments and Results}

We have generated 10 training traces of each type of behavior (good and bad) using the simulator explained in
Section~\ref{sec:simulator}. We have experimented using a higher number of training traces \change{and also generating
  an unbalanced training set where the good traces outnumber by a large margin the bad traces, as it is the case in real
  AML investigation. We observed that if a small number of training traces represent the prototypical traces of each
  class, $k$NN approaches can obtain good accuracy even in the unbalanced case.} The traces were synthetically generated
since there is no other available dataset that includes real data, apart from some transaction-based simulators. Since we can handle richer representation models than just transactional data,
these other datasets were of limited use in our case.
For evaluation, we have generated 20 new test traces that are randomly sampled from the two types of behavior. At each
step in the test traces, we use the classifier to predict the class of the new trace. We report the accuracy of the
prediction at the end of each test case, as well as how many observations, in average, the classifier needed in order to
generate the final decision (whether it was the correct or incorrect one). We have used $k=1$ for the experiments, given
that we already obtained good results. \change{We tried other values, and results were equivalent.}

\no{We first used \cabbot with traces generated by AMLSim. We used their generated data with 1K accounts and 10K
  transactions.\footnote{Available at \url{https://github.com/IBM/AMLSim}.} We converted the generated data into \cabbot
  input traces and randomly selected traces for training or test for both kinds of behavior. The only observable actions
  generated by that simulator were related to {\tt create-account} (implicitly defined in the accounts database that
  AMLSim generates) and {\tt move-funds}.
  The simulator does not generate the corresponding state representation, so we added the state changes related to these
  two actions. In the case of opening an account, we added the initial balance and the account owner and country. In the
  case of moving funds from one account to another, we added to the state the final balance, the amount of the
  transaction, and the origin and destination. The resulting accuracy is 0.5 if we use $d_a$, $d_\Delta$, and $d_g$ and
  0.55 if we use $d_r$. In the case of $d_a$ and $d_\Delta$, they classified all test instances as good behavior. The
  other two distance metrics varied the prediction, though they did not generate better than random.}

\subsection{Observability Models and Distances}

Tables~\ref{tab:observability-similarity} and~\ref{tab:accuracy} depict the effect of different observability models (rows) in combination with different distance functions (columns); i.e. $d_a$, action-based, $d_\Delta$, state difference, $d_g$, $n$-grams, and $d_r$, relational. We have used as observability models: full
($\observer$ can observe all actions and literals), bank (it can only observe the literals/actions related to
information that is provided by $\planning$ to $\observer$), network (it can
only observe from the bank model, data related to the relations of $\planning$ with other actors - customers or companies), transactions
(it can only observe from the bank model transaction-based information) and two more explained later.  Since we are performing
on-line classification, the values in Table~\ref{tab:observability-similarity} reflect the average number of pairs
action/state that $\observer$ had to observe before it made the final classification. Table~\ref{tab:accuracy} shows the
accuracy at the end of the trace.

\begin{table}[hbt]
\centering
\begin{tabularx}{\textwidth}{l *{4}{Z}}
  \toprule
    \multicolumn{1}{c}{\bf Observability} & \multicolumn{4}{c}{\bf Distance function}\\
    \multicolumn{1}{c}{\bf model} & \multicolumn{1}{r}{\bf $d_\Delta$} & \multicolumn{1}{r}{\bf $d_a$}
    & \multicolumn{1}{r}{\bf $d_g$} & \multicolumn{1}{r}{\bf $d_r$}\\ \midrule
    full & 0.50 & 0.50 & 9.0 & 0.6\\
    bank & 2.00 & 3.50 & 5.6 & 1.0\\
    network & 2.20 & 3.70 & 3.55 & 4.3\\
    transactions & 7.75 & 14.65 & 14.5 & 5.5\\
    no-companies & 6.05 & 15.70 & 18.4 & 2.6\\
    limited & 9.70 & 28.00 & 13.2 & 1.6\\ \bottomrule
  \end{tabularx}
  \caption{Average number of observations before making the final classification decision when varying the observable
    part of the model and the distance function using $k$NN. Columns represent the  distance functions: $d_a$, action-based; $d_\Delta$, state difference; $d_g$ $n$-grams; and $d_r$, relational.}
  \label{tab:observability-similarity}
\end{table}

We conclude that the more information is observable by $\observer$, the faster the right classification is made. In the
case of providing full information - the unrealistic case where the financial institution can observe all predicates and
actions -, it converges very fast to the right decision. After only one step, it commits to the right decision (perfect
accuracy) since it sees all the information, including whether someone is a criminal. In case of only observing
transactions data, it takes more time to converge than using only network information, since network data (opening
accounts, being part of companies, \ldots) is observed before customers start making transactions. Another observation
is that using the delta-distance provides better results than using actions-distance. This is expected as information on
the states is more diverse between the two types of behavior than the information on actions. States contain more knowledge than just what appears in actions' names and parameters. Also, given two consecutive states, we can infer the effects of the corresponding action, providing more information than just the name and parameters.

\begin{table}[hbt]
\centering
\begin{tabularx}{\textwidth}{l *{4}{Z}}
  \toprule
    \multicolumn{1}{c}{\bf Observability} & \multicolumn{4}{c}{\bf Distance function}\\
    \multicolumn{1}{c}{\bf model} & \multicolumn{1}{r}{\bf $d_\Delta$} & \multicolumn{1}{r}{\bf $d_a$}
    & \multicolumn{1}{r}{\bf $d_g$} & \multicolumn{1}{r}{\bf $d_r$}\\ \midrule
    full & 100 & 100 & 100 & 100\\
    bank & 100 & 100 & 100 & 100\\
    network & 100 & 100 & 100 & 100\\
    transactions & 100 & 100 & 90 & 100\\
    no-companies & 100 & 85 & 100 & 100\\
    limited & 100 & 65 & 95 & 100\\ \bottomrule
  \end{tabularx}
  \caption{Accuracy when varying the observable part of the model and the distance function using $k$NN. The columns represent the same distances as in Table~\ref{tab:observability-similarity}.}
  \label{tab:accuracy}
\end{table}

In most cases, the accuracy was perfect (100\%), so all test traces were correctly classified.  Even if we were very
careful on making the initial information and the actions taken by both kinds of behavior equal, \cabbot was able to
detect unintended differences in the traces. For instance, in the case of criminals, they create companies while we did
not implement that option for standard customers. To test the hypothesis that this provided an advantage to the network
based observability, we could have included those actions for regular customers. We report on that option in the next section. Instead, we created a new observability
type, no-companies, where $\observer$ could observe the same information as in the bank observability, except for any
company related information, such as predicate {\tt member-of} or action {\tt
  set-ownership-account}. Table~\ref{tab:observability-similarity} shows that in that case the performance is close to
that of only using transactions. 

Another example of the differences \cabbot found between the two kinds of behavior
relates to money withdrawals that were used by criminals while they were not used by regular customers or using digital
currency. So, we created another observability model, named limited, where $\observer$ could not observe the companies
related information, the withdrawals nor the operations with digital currency. In that case, it affects actions {\tt
  cash-out}, {\tt integration-cash-out}, {\tt digital-deposit}, {\tt placement-digital} and {\tt buy-digital}. The table
shows the results which obviously are worse than the other observation models. Also, in the case of using
actions-distance, the accuracy of these two last observation models dropped to 85\% and 65\%, respectively.
In relation to the distance functions defined, the accuracy is similar in most cases, but the time it takes them to
converge to the right classification varies from the simplest ones, $d_a$ and $d_\Delta$ to the most ellaborated one,
$d_r$. Using this last distance function, it needs very few examples to make the right decision.

\subsection{Traces Length and Goals Probability}

In the previous experiment we fixed a length of the trace to be 50. The second experiment aims at analyzing the effect
of the length of traces (simulation horizon) in the accuracy, fixing the observation model to bank and the distance
metric to $d_r$.
Again, we obtained a perfect accuracy starting with traces of length 5, given that the creation of companies was
performed in the early stages.
Since these results depend on how often a customer performs actions, we changed the probability of new goals being
generated in a simulation step. This probability affects how often a customer performs actions as explained in
Section~\ref{sec:simulator}. We varied that probability and checked against different horizons. Results are shown in
Table~\ref{tab:horizon-prob-goal}. We can observe that if the probability of a goal appearing in a given step decreases,
the accuracy also decreases, since less observations are made by $\observer$. If the trace is short or the probability
of a goal appearing is small, there is less space for \cabbot to detect bad/good behavior. So, it becomes equivalent to
a random decision. For instance, when probability is 0.01, a goal will only appear once every 100 steps, so the
classification will be based on no information.

\begin{table}[hbt]
\centering
\begin{tabularx}{\textwidth}{ *{7}{Z}}
  \toprule
    & \multicolumn{6}{c}{\bf Length of the observed trace}\\
    \multicolumn{1}{c}{\bf Prob. goal} & \multicolumn{1}{r}{\bf 1} & \multicolumn{1}{r}{\bf 5} & 
    \multicolumn{1}{r}{\bf 10} & \multicolumn{1}{r}{\bf 20} & \multicolumn{1}{r}{\bf 50}  &
    \multicolumn{1}{r}{\bf 100}\\ \midrule
    1.0 & 80 & 100 & 100 & 100 & 100 & 100\\
    0.8 & 60 & 100 & 100 & 100 & 100 & 100\\
    0.5 & 50 & 70 & 100 & 100 & 100 & 100\\
    0.2 & 60 & 55 & 75 & 100 & 100 & 100\\
    0.1 & 60 & 60 & 70 & 80 & 95 & 100\\
    0.05 & 70 & 45 & 50 & 70 & 80 & 100\\
    0.01 & 40 & 65 & 50 & 50 & 50 & 65\\ \bottomrule
  \end{tabularx}
  \caption{Classification accuracy when varying the traces length and the probability a goal appears at a given
    time step.}
  \label{tab:horizon-prob-goal}
\end{table}

\subsection{Comparison against a Non-Relational Representation}

The aim of this experiment is to improve the variety of traces generated by the simulator. First, we allowed standard
customers to create companies, making the traces much more diverse. Second, we wanted to compare against a learning
technique used in other works and suitable in terms of explainability for the purposes of AML investigation.
Table~\ref{tab:ribl-flat} shows a comparison of \cabbot and a decision tree classifier. In order to use the decision
tree, we had to convert the traces to an equivalent attribute-value representation. For all training traces, we
generated training examples by observing the first action-state pair, the first two action-state pairs, and so on until
the length of the trace. For each action-state pair, we created standard attributes used by other works for the two
partial observability models (under the bank and full models, we could observe all these attributes). Examples are
average, min and max values of the previous transactions of each type (e.g. wires, or deposits), balance of accounts or
number of connected accounts.

\begin{table*}[hbt]
\centering
\begin{tabularx}{\textwidth}{l *{5}{Y}}
  \toprule
      & \multicolumn{5}{c}{\bf Length of the observed trace}\\
    \multicolumn{1}{c}{\bf Observability} & \multicolumn{1}{c}{\bf 10} & \multicolumn{1}{c}{\bf 20}
     & \multicolumn{1}{c}{\bf 50} & \multicolumn{1}{c}{\bf 100} & \multicolumn{1}{c}{\bf 350}\\
    \multicolumn{1}{c}{\bf model} & \multicolumn{5}{c}{\bf {\sc Decision tree}}\\ \midrule
    full & 75 (2.7) & 90 (6.8) & 95 (10.4) & 95 (16.9) & 95 (24.0)\\
     bank & 75 (2.7) & 90 (6.8) & 90 (8.3) & 95 (15.7) & 95 (24.1) \\
     transactions & 60 (0.6) & 90 (6.0) & 90 (8.3) & 90 (8.1) & 85 (39.1) \\
     network & 55 (0.0) & 75 (2.4) & 85 (5.9) & 85 (5.9) & 85 (5.9) \\
     \midrule
      & \multicolumn{5}{c}{\bf \cabbot}\\
    full & 100 (1.1) & 100 (1.1) & 100 (1.1) & 100 (1.1) & 100 (1.1)\\
     bank & 100 (2.3) & 100 (2.3) & 100 (2.3) & 100 (2.3) & 100 (2.3)\\
     transactions & 90 (2.2) & 80 (7.7) & 100 (9.8) & 90 (16.8) & 90 (20.2)\\
     network & 80 (4.9) & 80 (4.9) & 80 (4.9) & 80 (11.3) & 85 (13.9)\\
     \bottomrule
  \end{tabularx}
  \caption{Accuracy and average number of observations before making the final classification decision (in parenthesis) when varying the
    observable part of the model and the length of the observed trace.}
  \label{tab:ribl-flat}
\end{table*}

We varied the observability models and the traces' length. In general, \cabbot obtains better performance both in
accuracy and number of observations needed to obtain the correct classification. We can also see that the full and bank
observability models obtain very good results, but performance degrades when using transactions or network.

\section{Generation of Synthetic Behavior}
\label{sec:simulator}

Available AML-related datasets mostly only include transactional data. We have built a simulator that uses automated planning to generate traces that provide a richer and more realistic representation of the information a financial institution can observe about its customers and their financial transactions. The simulator uses automated planning to generate the traces. The only other two simulators specific to money laundering we are aware of are
AMLSim~\cite{weber2018scalable},\footnote{\url{https://github.com/IBM/AMLSim}} and~\cite{lopezrojas}. Both mostly focused on transaction data.
Currently, AMLSim can generate some money laundering behaviors (fan-in, fan-out and cyclic).
Instead, our simulator allows for simulating richer money laundering behavior by allowing abnormal transfer pricing, or interleaved standard behavior.
Also, our simulator incorporates a richer network structure,
such as customers being companies, owned by networks
of people. 

\begin{figure}[hbt]
\placefigure{0.4}{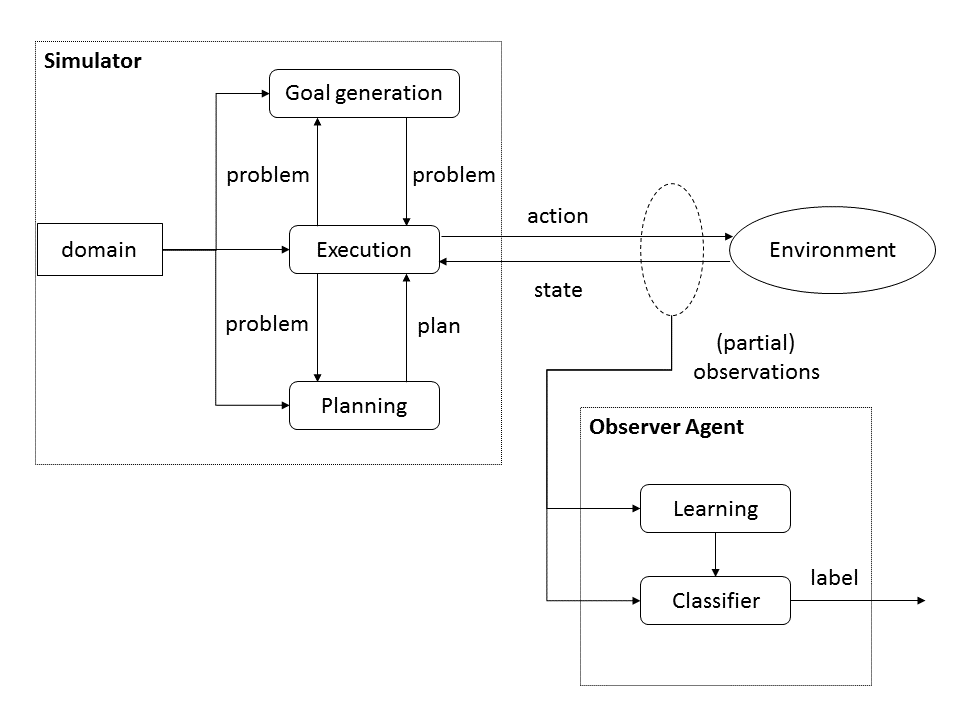}
\caption{High level view of the simulator.}
\label{fig:arch}
\end{figure}

Figure~\ref{fig:arch} shows
an outline of the simulator (corresponding to $\planning$) and the observer (corresponding to $\observer$.
$\planning$ takes actions in the environment by using a rich
reasoning model that includes planning, execution, monitoring and goal generation as explained below.

\subsection{Domain Model}

We model the domain with PDDL (Planning Domain Description Language), that allows for a compact representation of
planning tasks~\cite{PDDL}. We define a common domain model for both behaviors (standard and criminal). It includes: a hierarchy of types (e.g. account, company, or customer); a set of predicates and functions (Tables~\ref{tab:predicates} and~\ref{tab:functions}); and a set of actions (Table~\ref{tab:actions}).

Different planning problems can be defined within a domain. They consist of: (1) a set of objects, such as customers,
accounts, companies, and transactions; (2) an initial state that, in our case, is the same for both kinds of behavior except
that it contains information on a customer being a criminal for bad behaviors; and (3) a set of goals, that is initially
empty as they will be dynamically generated by the goals generation component.

We have modeled some examples of known behavior for money
laundering~\cite{Irwin12}. Standard money laundering is composed of placement, layering and
integration. Placement consists of introducing the money with illicit origin into the financial system. We have
implemented two different ways of performing placement: depositing money directly into bank accounts, or moving digital money to standard accounts. Layering consists on moving {\it placed} money
into other accounts to make tracing the origin/destination of money difficult. 
Integration consists on using that money for standard operations. We have implemented three
integration strategies: withdrawal, paying bills, or international money transfer. All these decisions
are made randomly according to some probability distributions. 

\subsection{Execution}

At each simulation step, the execution component calls the planning component if there is a reason for (re)planning. Reasons for replanning are:
the new state is not the expected one; or goal generation has returned new goals.  If there is
a plan in execution, it simulates its execution in the environment.
Our simulator includes the possibility of defining deterministic and non-deterministic execution of actions, and
the appearance of exogenous events. At each step, the execution component calls goal generation for changes in the goals or partial descriptions of states, as explained below.

The interaction with the environment also generates a trace of observations that will be used for training and test
of $\observer$ learning component. The trace contains a sequence of actions and states from the $\observer$ viewpoint. Hence, the execution applies a filter on both so that it includes only its observable
elements  in the trace.
Each simulation finishes after a predefined number of simulation steps (horizon) that is a parameter, or after a plan
was not found in two consecutive steps in a given time bound. We set the time bound with a low value (10 seconds), since
this is enough in most cases.

\subsection{Goal Generation}

This component allows agents to generate realistic behavior whose goals evolve over time depending on the current state
of the environment. It takes as input the current problem description (state, goals and instances) and returns a new
problem description. The first obvious effect of this module is to change goals. In order to do so, we have defined two
kinds of behavior by changing the goals of each type. In the case of persons doing money
laundering, the simulator will dynamically generate goals corresponding to a pure bad behavior, such as committing a
crime, or laundering money. But, the simulator will interleave these bad behavior goals with standard customer goals, so
that the task of deciding whether some trace belongs to a bad behavior is not easy to detect. In the case of standard
customer goals, the simulator would generate goals such as owning a house (or cheaper kinds of products or services), working for a company, creating a company, or making payments to an utility company.
The generation of goals for both kinds of behavior is guided by some probability distributions that allow to easily change the types of traces generated.

This component can also change the state and instances. This is useful for generating new components of the state
on-line. As an example, we do not want to include initially information about all transactions to be performed by
$\planning$ during the complete simulation period. Instead, the Goal generation allows the simulator to define new objects or state components as needed. So, if it generates a goal of buying a product, it could generate a new customer -- the seller--,
her account, the product to be bought, and all the associated information in the state.

\subsection{Planning}

Planners take as input a domain and problem description in PDDL, and return a plan that solves the
corresponding planning task. 
In principle, we could have used any PDDL complaint planner. However, we make extensive use of numeric variables (using PDDL functions). So, we are restricted to planners that can reason with numeric preconditions and effects. We are using {\sc cbp}~\cite{CBP}.

\subsection{Experiments on Behavior Novelty}

As a final experiment, we wanted to test the hypothesis that the learning system would be able to still correctly identify new unseen behavior, which would lead to a great advantage to AML investigation. In order to test the identification of novel behavior, we generated 10 training instances corresponding to a specific money laundering behavior (placement with cash deposits,  usually called structuring) and also 10 training instances of standard customer behavior. Then, we generated 20 test instances randomly selecting good and bad behavior. However, now the bad behavior used placement with digital money. So, we were trying to test the ability of \cabbot of correctly classifying unseen behavior. We used long traces of 350 steps. The result is that it correctly classified all test instances (100\% accuracy). But it took it more observations than before to detect it; from an average of 2.75 when it saw both kinds of placement in the training instances, to 8.3 when it only saw cash deposits in the training instances and the test where using digital. In the case of the decision tree, it also had a 100\% accuracy, but the average number of observations that it required to converge to the right decision was 50.95. When we reversed the training behavior (digital placement) and test behavior (cash deposits), we obtained equivalent results. These results are very encouraging from the point of real investigation, since the system is able to correctly classify unseen behavior (from a different probability distribution).

\section{Conclusions}

We have presented three main contributions. The first one consists of a new way of modeling the AML task as
classification of relational-based behavior traces. The models of observed behaviors are richer that usual ones based on
transactions only. They also allow to easily create different observation models and evaluate the impact of using
them. The second contribution is \cabbot, a relational learning technique that takes a set of training traces of other
agents's behavior and can classify later traces. We have used some known relational distances and have proposed a new
one that explicitly considers the structure of the input examples (traces). The third contribution is a simulator that
generates synthetic behaviors where agents can dynamically change their goals, and therefore their plans. The execution
of those plans leaves a behavior trace that can be used by the learning system to classify new traces.  The diversity of
generated traces can be used to challenge the learning system, as well as it can be exploited by financial institutions
to pro-actively study new kinds of behavior.  Experimental results show that \cabbot can successfully learn to classify
new traces in the presence of different observability models. In future work, we would like to connect the learning
component to real data, and automate the process of AML using the output of its decisions.

\section*{Acknowledgements}

\change{This paper was prepared for information purposes by the Artificial Intelligence Research group of JPMorgan Chase
  \& Co. and its affiliates (“JP Morgan”), and is not a product of the Research Department of JP Morgan.  JP Morgan
  makes no representation and warranty whatsoever and disclaims all liability, for the completeness, accuracy or
  reliability of the information contained herein.  This document is not intended as investment research or investment
  advice, or a recommendation, offer or solicitation for the purchase or sale of any security, financial instrument,
  financial product or service, or to be used in any way for evaluating the merits of participating in any transaction,
  and shall not constitute a solicitation under any jurisdiction or to any person, if such solicitation under such
  jurisdiction or to such person would be unlawful. The authors thank Alice Mccourt for her useful
  revision of the paper.}
 

\bibliographystyle{named}

\end{document}